\documentclass[conference]{IEEEtran}

\usepackage{amsmath}
\usepackage{amsthm}
\newtheorem{theorem}{Theorem}

\newtheorem{definition}{Definition}

\usepackage{graphicx} 
\usepackage{algorithm}
\usepackage{algorithmic}
\usepackage{amssymb}
\usepackage{tabularx}
\usepackage{booktabs}
\usepackage{url} 
\newcolumntype{L}[1]{>{\raggedright\arraybackslash}p{#1}}
\newcolumntype{C}[1]{>{\centering\arraybackslash}p{#1}}
\newcolumntype{R}[1]{>{\raggedleft\arraybackslash}p{#1}}

\usepackage{array} % for better columns
\usepackage{comment}

\title{Anomaly Detection in Complex Dynamical Systems: A Systematic Framework Using Embedding Theory and Physics-Inspired Consistency}

\author{

\IEEEauthorblockN{Michael Somma\IEEEauthorrefmark{1}\IEEEauthorrefmark{3}, Thomas Gallien\IEEEauthorrefmark{2}\IEEEauthorrefmark{4}, Branka Stojanović\IEEEauthorrefmark{1}}

\IEEEauthorblockA{
\IEEEauthorrefmark{1}JOANNEUM RESEARCH Forschungsgesellschaft mbH,\\ 
DIGITAL – Institute for Digital Technologies,\\
Steyrergasse 17, Graz, Austria, 8010\\
Email: \texttt{michael.somma@joanneum.at},\\\texttt{branka.stojanovic@joanneum.at} 
}

\IEEEauthorblockA{
\IEEEauthorrefmark{2}JOANNEUM RESEARCH Forschungsgesellschaft mbH,\\ 
ROBOTICS – Institute for Robotics and Flexible Production,\\
Lakeside B13b, Klagenfurt, Austria, 9020\\
Email: \texttt{thomas.gallien@joanneum.at} 
}

\IEEEauthorblockA{
\IEEEauthorrefmark{3}TU Graz, Institute for Technical Informatics, Inffeldgasse 16/I, Graz, Austria, 8010 
}

\IEEEauthorblockA{
\IEEEauthorrefmark{4}AI AUSTRIA, RL Community, Wollzeile 24/12, Vienna, Austria, 1010
}

}

\begin{document}

\maketitle

\begin{abstract}
Anomaly detection in \textit{complex dynamical systems} is essential for ensuring reliability, safety, and efficiency in industrial and cyber-physical infrastructures. \textit{Predictive maintenance} helps prevent costly failures, while \textit{cybersecurity monitoring} has become critical as digitized systems face growing threats. Many of these systems exhibit oscillatory behaviors and bounded motion, requiring anomaly detection methods that capture structured temporal dependencies while adhering to physical consistency principles. In this work, we propose a \textit{system-theoretic approach} to anomaly detection, grounded in \textit{classical embedding theory} and \textit{physics-inspired consistency principles}. We build upon the \textit{Fractal Whitney Embedding Prevalence Theorem} that extends traditional embedding techniques to complex system dynamics. Additionally, we introduce state-derivative pairs as an embedding strategy to capture system evolution. To enforce temporal coherence, we develop a \textbf{Temporal Differential Consistency Autoencoder (TDC-AE)}, incorporating a TDC-Loss that aligns the approximated derivatives of latent variables with their dynamic representations. We evaluate our method on two subsets (FD001, FD003) of the \textbf{C-MAPSS} dataset, a benchmark for turbofan engine degradation. TDC-AE machtes LSTMs and outperforms Transformers while achieving a nearly \textbf{100x reduction in MAC operations}, making it particularly suited for \textit{lightweight edge computing}. Our findings support the hypothesis that anomalies disrupt stable system dynamics, providing a robust signal for anomaly detection.  
\end{abstract}

\vspace{0.2cm}
%keywords with commas.

\section{Introduction}
Anomaly detection in complex physical dynamical systems is a critical research area as industrial and engineered systems become more sophisticated. Identifying deviations from expected behavior is essential for ensuring reliability, safety, and efficiency. This is particularly relevant in predictive maintenance and cybersecurity monitoring. In industrial systems and rotating machinery, early fault detection helps prevent costly failures and downtime \cite{nunes_challenges_2023}. 
Meanwhile, as critical infrastructures, such as power grids and water distribution systems, become increasingly digitized, the risk of cyber threats has grown significantly \cite{tuptuk_systematic_2021,riggs_impact_2023}. Recent attacks \cite{veolia2024incident,bajak_hackers_2023} on cyber-physical systems highlight the need for robust monitoring techniques to detect both malicious intrusions and system failures. Ensuring the security and stability of these dynamical systems requires adaptive anomaly detection methods capable of addressing evolving threats and operational challenges.

Many complex dynamical systems exhibit oscillatory behaviors and bounded motion, fundamental characteristics of both natural and engineered processes. The study of such systems intersects with two key fields: \textbf{time-series modeling} and the \textbf{incorporation of physical laws}. Given that many physical systems display structured temporal dependencies, effective modeling requires methods that capture correlations across time. Simultaneously, classical physics, which governs tangible objects and engineered systems, is largely defined by causal and deterministic principles, often described through differential equations and conservation laws. This study aims to bridge time-series modeling with physics-inspired approaches to develop more effective and sustainable anomaly detection methods.

\section{Related Work}
\subsection{Time-Series Modeling}
Models like \textit{LSTMs} and \textit{Transformers} have been successfully applied to time-series anomaly detection in fields such as mechanical engineered system, aerospace, and industrial monitoring \cite{lachekhab_lstm-autoencoder_2024,liu_deep_2023,wei_lstm-autoencoder-based_2023}. While these methods effectively capture long-range dependencies and irregular temporal patterns, they are computationally expensive. For example, evaluated on common benchmark datasets, sequence lengths up to 500 time steps were required to achieve competitive performance \cite{mahmoud_ae-lstm_2022}. The high memory requirements often exceed the constraints of typical MCU-level devices, making real-time deployment challenging. For example, a typical LSTM with 500 time steps and 50 hidden units already surpasses the available memory of common low-power devices, limiting its practical applicability. Additionally, the sequential nature of RNNs restricts parallelization, further increasing computational cost \cite{rezk_recurrent_2020}. These challenges highlight the need for alternative approaches that balance computational efficiency with robust anomaly detection, aligning with broader goals of sustainability and practical deployability. \par
\subsection{Physics-Informed Methods}
In many applications, incorporating domain knowledge can help reduce computational demands. Physics-informed neural networks (PINNs) embed physical laws directly into neural networks, enabling them to leverage known system dynamics \cite{cai_physics-informed_2021,raissi_physics-informed_2019,raissi_hidden_2020}. However, applying PINNs to complex dynamical physical systems remains challenging. These systems involve numerous interdependent physical principles, and explicitly modeling them within a neural network would require domain expertise, making large-scale applications impractical \cite{karniadakis_physics-informed_2021,wang_when_2022}.

\subsection{Benchmark Use Case: Turbofan Aeroengine Degradation}
We use the C-MAPSS dataset as a benchmark for studying complex dynamical systems due to its realistic representation of turbofan aeroengine degradation. Aeroengines are complex dynamical systems governed by nonlinear, time-dependent interactions of physical processes, making them an ideal test case for evaluating anomaly detection methods in real-world settings.\par

In the literature, anomaly detection for the C-MAPSS dataset generally follows two main approaches. The first focuses on fleet-level anomaly detection, where engines with shorter lifetimes are classified as abnormal based on their total life cycles \cite{jakubowski_anomaly_2022,yildirim_enhancing_2024}. This method aims to distinguish early failures from normal operating conditions at a system-wide level.

The second approach considers individual engine degradation, defining anomalies based on a 60/40 time-based split \cite{bataineh_autoencoder_2020,zhu_anomaly_2024}. In this setup, the first 60\% of an engine’s life is labeled as normal, while the final 40\% is considered abnormal. The dataset is then divided into train/test subsets, and models are evaluated based on their ability to classify each time step accordingly.

Since we aim to develop methods that capture system dynamics, we consider the second approach a more suitable benchmark, as it focuses on time-dependent degradation rather than static fleet-level classification.

Previous work on anomaly detection in this setting has explored various deep learning models. One study employed an LSTM-based approach \cite{zhu_anomaly_2024}, leveraging recurrent structures to model time dependencies. Another approach used a Vanilla Autoencoder (AE) without explicit temporal modeling \cite{bataineh_autoencoder_2020}. More recent research has investigated Transformer-based models, which excel at capturing long-range dependencies but introduce high computational costs and require extensive training data due to their large parameter space \cite{inproceedings,liu_deep_2023}. Moreover, both Transformer studies formulated the problem as a multiclass classification task, with one using a slightly different dataset and the other applying the method directly to C-MAPSS. However, defining well-separated fault categories in a complex dynamical system is challenging in practice, as non-trivial interactions between multiple physical components create highly unpredictable behaviors. As system dynamics grow more intricate, these interactions become even less predictable, further complicating precise fault categorization.

\section{Theoretical Foundation \& Methods}

\subsection{Proposed Approach \& Motivation}
We propose a data-driven anomaly detection framework grounded in physics consistency and embedding theory, without explicitly enforcing known physical laws. Our central premise is that effective anomaly detection in dynamical systems requires understanding their intrinsic dynamics.

We model system behavior using state-derivative pairs, a natural representation for dynamical systems. From observed measurements, we construct a mapping into a state-space-like manifold that approximates the system’s latent dynamics. Under normal operating conditions, we hypothesize that the system evolves within a well-defined, compact set of dynamic states, and that anomalies manifest as deviations from this set.

Consider, for example, a damped-driven pendulum. Its phase space, defined by angular displacement $\theta$ and its derivative $\dot{\theta}$, initially confines the system to a bounded, oscillatory regime. However, under drift, the trajectory may exit this region, signaling a departure from regular dynamics. This behavior exemplifies how anomalies manifest as transitions from compact, predictable trajectories to unbounded or inconsistent dynamics (see Fig.\ref{fig:CMPSS_pendulum_theta_phasespac}, implementation details in Appendix~\ref{sec:definitions_implementation}).

To justify this assumption, the learned mapping must be \emph{locally invertible and injective} in the normal regime, ensuring that nearby observations correspond to unique and consistent states in the latent space.

Embedding theory provides guidance for achieving injectivity by relating the geometric complexity of the measurement manifold to a lower bound on the embedding dimension required for faithful reconstruction of system dynamics.

We assume that anomalies introduce additional \textbf{dynamic complexity}, which increases the effective \textbf{geometric complexity} of the measurement space. Our goal is to construct an embedding whose geometric complexity is \emph{matched to the dynamic complexity of normal operation}, sufficient to represent typical behavior, but not overparameterized to absorb arbitrary deviations.

When an anomaly occurs, the resulting increase in complexity exceeds the representational capacity of the embedding, leading to an inconsistent or distorted description of the latent dynamics. We exploit this mismatch as a signal for anomaly detection.

To validate our assumptions, we use numerical methods to verify that the Jacobian of the learned mapping is full-rank (for invertibility) and to empirically assess local injectivity.

\subsection{Embedding Complex System Dynamics into a Latent Space}

We begin by formalizing the mathematical foundation for embedding high-dimensional dynamical systems into a reduced representation. 
Consider a \textbf{complex bounded dynamical system} with state variables $x \in \mathbb{R}^{m}$, where $m$ represents the dimension of the \textbf{physical system} and is typically much larger than the number of available measurements $( m \gg k$).

At the \textbf{measurement level}, we access observables through a measurement function:
\begin{equation}
\mu: \mathbb{R}^{m} \to \mathbb{R}^{k},
\end{equation} \label{eq:Mu}
where \(\mu \) extracts a lower-dimensional set of measurements \( y = \mu(x) \), with \( y \in \mathbb{R}^{k} \), forming the input to the \textbf{autoencoder}.

The \textbf{encoding level} is defined by:
\begin{equation}
\mathcal{E}: \mathbb{R}^{k} \to \mathbb{R}^{n},
\end{equation}\label{eq:Epsilon}

where \( \mathcal{E} \) maps the measurement space to an \textbf{embedding space} of dimension \( n \).

Embedding measurements into a lower-dimensional space, as described above, is a well-studied problem in \textbf{classical embedology}. Embedding theorems, such as those by Whitney, provide conditions under which one can construct injective maps from measurement data to a lower-dimensional space. Specifically, these theorems relate the dimensionality of the manifold covered by the measurements to the ambient space in which the measurements reside.

Ideally, systems in stable operation evolve on smooth manifolds. However, real-world dynamics often lie on more complex, non-smooth sets due to measurement noise and the inherently intricate behaviors required by physical systems.

To generalize the idea of dimension for such cases, we use the \textit{box-counting dimension}, which extends naturally to non-smooth subsets. 
For a heuristic and accessible explanation of this concept and a formal definition, see Appendix~\ref{sec:definitions_implementation}.)

The \textit{Fractal Whitney Embedding Prevalence Theorem}~\cite{sauer_embedology_1991} (see Appendix~\ref{sec:definitions_implementation}) states that if the measurement data lie in a compact subset of the measurement space, and if the box-counting dimension of this set is $d$, then an embedding into $\mathbb{R}^n$ with $n > 2d$ is generically injective. This provides a practical criterion: estimate the box-counting dimension of the measurements and choose an embedding dimension that exceeds twice this value to obtain an injective mapping with high probability.

Having established the role of embedding theory, we now turn to \textbf{physical consistency} part. We draw inspiration from simple dynamical systems and propose using \textit{state-derivative pairs} as the embedding. This choice is motivated by two principles. First, physical laws typically encode \textit{causal relationships} between a system's state and its derivative, making this representation natural for dynamical processes. Second, state-derivative pairs help \textit{reduce redundancy} by promoting linear independence between embedding dimensions.

To be effective, such embeddings must not only be injective, but also preserve a \textit{consistent differential structure}. That is, small changes in the latent space should induce meaningful, non-degenerate changes in the measurement space. A full-rank Jacobian is a necessary and sufficient condition to ensure this well-behaved, locally reversible mapping~\cite{lee_introduction_2012}.
\begin{figure}
    \centering
    \includegraphics[width=0.9\linewidth]{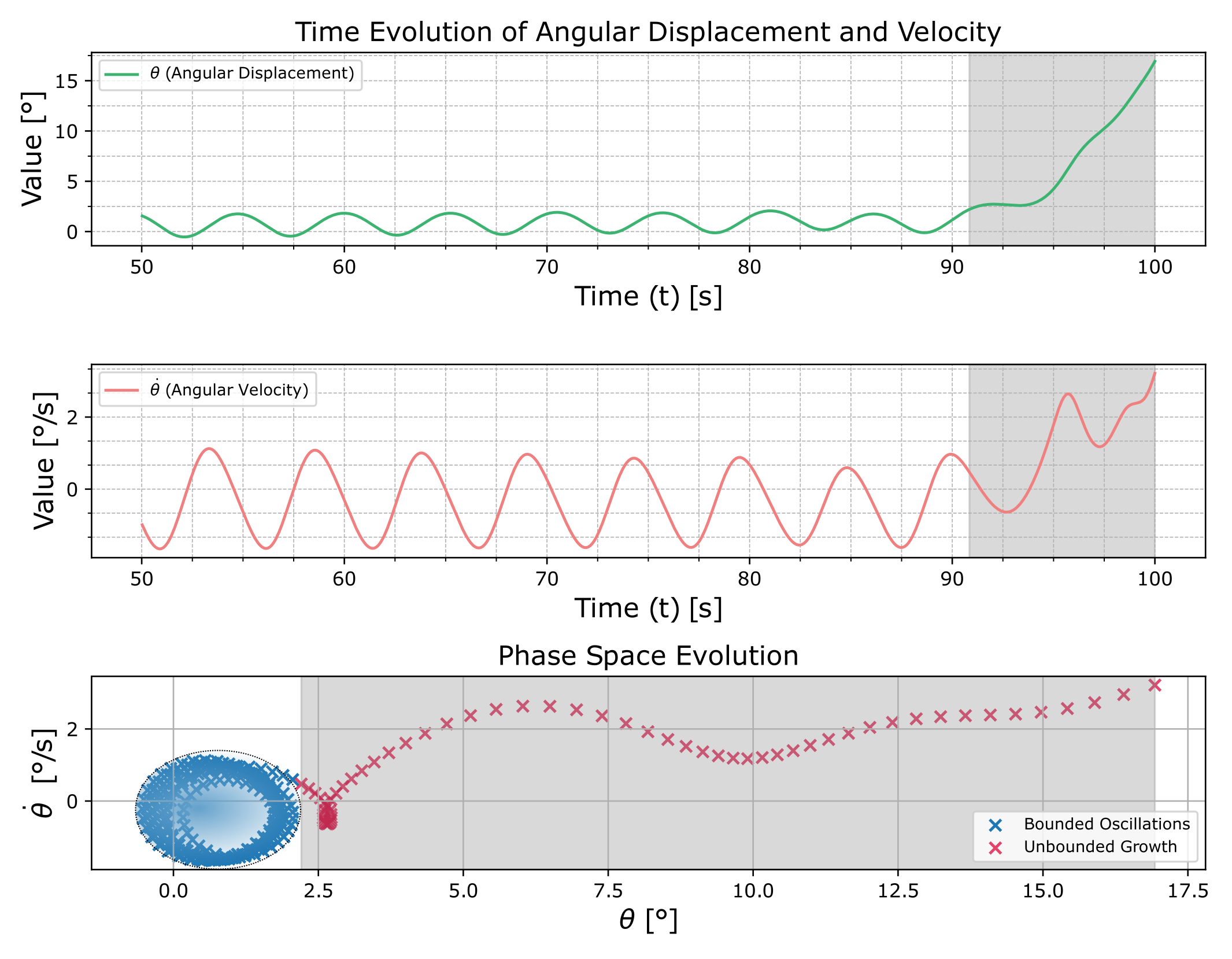}
        \caption{Phase plot of a damped forced motion, showing $\theta$ and $\dot{\theta}$. The \textbf{blue trajectory} represents the stable phase space evolution, while the \textbf{red trajectory} highlights an unstable deviation from the stable state.}

    \label{fig:CMPSS_pendulum_theta_phasespac}
\end{figure}

We use intrinsic dimensionality (ID) as a proxy for the boxcounting dimension. We applied a per-engine estimation procedure using the Two-Nearest-Neighbor (Two-NN) method \cite{facco_estimating_2017}. This estimator is particularly well suited for our setting, where each engine provides a small number of samples that are assumed to lie on a low-dimensional manifold.

To empirically assess whether our mapping preserves injectivity, we compute the ratio of pairwise distances in the latent space \( Z \) to those in the input space \( X \), i.e., \( \|\mathcal{E}(x_1) -  \mathcal{E}(x_2)\| / \|x_1 - x_2\| \), for input pairs satisfying \( \|x_1 - x_2\| > \delta \) (with \( \delta = 10^{-6} \)). A minimum ratio below \(10^{-5}\) indicates potential collapse and loss of injectivity. This threshold is chosen based on standard floating-point precision limits and empirical practices in representation learning~\cite{Goodfellow-et-al-2016}. The method serves as a practical diagnostic for identifying when distinct inputs are mapped to nearly identical representations.

%%%%%%%%%%%%%%%%%%%%% HERE %%%%%%%%%%%%%%%%%%%%%%%%%%%%%%%%%%%%

\subsection{Application to Autoencoder-based Representations}
With the theoretical framework established, we select a stable, bounded dynamical system with accessible measurements. For this, we use the NASA \textbf{C-MAPSS dataset}, which provides time-series sensor data from aircraft engines.
\subsubsection{C-MAPSS Dataset}

The NASA Commercial Modular Aero-Propulsion System Simulation (C-MAPSS) dataset \cite{Saxena2008CMAPSS} is a widely used benchmark in prognostics and health management studies \cite{Saxena2008Damage,AlDulaimi2019,Benker2020}. 

The dataset consists of multivariate time series capturing engine operation under varying initial wear, manufacturing differences, and noisy sensor readings. Each engine starts in a normal state before gradually developing a fault.

While originally designed for Remaining Useful Life (RUL) estimation, the C-MAPSS dataset has become more popular lately in health management studies for \textbf{anomaly detection, fault diagnostics, and predictive maintenance}~\cite{jakubowski_anomaly_2022}. Its structured time-series format and realistic degradation patterns make it valuable for developing and testing machine learning models aimed at early fault detection and system health assessment.  

We use the \textbf{FD001} and \textbf{FD003} subsets of the C-MAPSS dataset, which differ in fault complexity: FD001 includes one fault mode, while FD003 includes two.
For anomaly detection, we follow a 60/40 split, where the first 60\% of cycles are labeled as normal and the remaining 40\% as anomalous, as suggested in similar studies~\cite{bataineh_autoencoder_2020,zhu_anomaly_2024}. A random 80/20 train-test split based on separate engines ensures robust evaluation.

The dataset includes sensor readings for temperature, pressure, fan and core speed, efficiency ratios, fuel flow, and coolant bleed parameters, providing insights into engine performance and degradation.

\subsubsection{Applying the Theoretical Foundation}
The NASA C-MAPSS dataset provides a measurement space of dimension \( k = 24\), representing sensor readings from the turbine system. To determine a suitable embedding dimension in accordance with Th.~\ref{th:fractal_emeddbing}, we establish the following assumptions about the system’s essential dynamics.

We assume that the measurement function in the C-MAPSS dataset retains sufficient information about the underlying system dynamics. Given the structured and high-resolution nature of the sensor data, the measurement space is taken to provide a reliable, non-degenerate representation of the turbine's evolving state---a standard assumption in physical modeling from observation data.

\paragraph{Intrinsic Dimensionality and Embedding.}
Using the Two-NN estimator as a proxy for the box-counting dimension, we obtain the following intrinsic dimension estimates for each dataset subset: F001: \( 4.93 \pm 0.75 \), and F003: \( 4.83 \pm 0.65 \). Interpreting these as fractal dimensions of the measurement space, and applying Theorem~\ref{th:fractal_emeddbing}, we require an embedding space of dimension greater than twice the fractal dimension. This yields lower bounds of \( 9.86 \) and \( 9.66 \), respectively. Rounding to the next even integer to accommodate both state and state-derivative pairs, we select embedding dimensions \( n = 10 \) for both F001 and F003.

Th.~\ref{th:fractal_emeddbing} assumes a smooth map and that it applies to "almost every" map. A \(\tanh\)-activated neural network encoder reasonably satisfies the smoothness assumption. The "almost every" condition applies to all smooth maps, whereas neural networks form a restricted subset. However, by universal approximation, a sufficiently large network can approximate such an embedding, making the theorem a strong and relevant practical guideline.

\paragraph{Rank Verification via SVD.}
To ensure that the reconstructed state matrix \( X \) captures the full dynamics, we verify its rank using singular value decomposition (SVD). The rank is determined by the number of non-zero singular values in \( \Sigma \); if all are non-zero, the matrix is full-rank~\cite{brunton_data-driven_2019}.

\subsubsection{Temporal Differential Consistency informed Autoencoder}

As a foundational step, we require a method to approximate the first-time derivative. A widely used approach is the \textit{central difference method}, which estimates the derivative of a function by computing the slope between two points symmetrically positioned around the point of interest. This method is often preferred over forward or backward differences due to its higher accuracy (\( O(\Delta t^2) \)), meaning that the approximation error decreases quadratically as \( \Delta t \) decreases \cite{chapra_numerical_2015}.

To accurately capture temporal dynamics, we incorporate the central difference method into the training framework of the latent space. The latent representations at the previous (\( t-1 \)) and next (\( t+1 \)) time steps are used to approximate the first-time derivative of the static latent variables (\( z \)) using the central difference formula. This derivative is then used as a target for the dynamic latent variables (\( \dot{z} \)), ensuring consistency with the central difference approximation, scaled by the time interval \( \Delta t \). A schematic representation of this approach is shown in Fig.~\ref{fig:TDC_AE_scheme} in Appendix~\ref{sec:figures}.

To incorporate temporal dynamics into the training process, we introduce a temporal differential consistency loss (TDC-Loss). This loss enforces consistency between the approximated time derivative of the static latent variables (\( z \)) and the output of the corresponding dynamic latent variables (\( \dot{z} \)). During training, the central difference method is used to estimate the time derivative of \( z \), which is then compared with \( \dot{z} \). The TDC loss serves as a regularizer for the standard reconstruction loss, ensuring that the autoencoder learns a latent representation that captures both state and derivative information. A compact version of the pseudocode for TDC-informed training is presented in Algo.~\ref{alg:TDC_informed_training} in Appendix~\ref{sec:Implementation}. We use a highly compact autoencoder with three hidden layers in both the encoder and decoder. For details on implementation and hardware, see Appendix~\ref{sec:Implementation}.

\subsubsection{Consistency Metrics}
We employ an embedding based on state-derivative pairs, where the derivative is approximated using the central difference method. While numerical differentiation is generally ill-posed, particularly in the presence of noise \cite{van_breugel_numerical_2020}, our approach does not seek a fully precise description of the system's dynamics. Instead, we aim for a causal approximation that reliably indicates when the system deviates from normal variability. By choosing a sufficiently large embedding dimension, we ensure that the system operates on a simple attractor geometry. This allows us to expect a nearly constant trendline for both the state and its derivative. To evaluate this, we introduce two metrics, which are detailed in the following paragraphs.\par

To ensure a balanced causal approximation between state and derivative, their variations must remain comparable. We define this using the ratio of their min-max scaled variances:

\begin{equation}
    \eta = \frac{\sigma_{\dot{x}_\text{scaled}}^2}{\sigma_{x_\text{scaled}}^2},
\end{equation}

where $\sigma^2$ denotes variance after min-max scaling. If $\eta \approx 1$, the variations are well-matched, making central difference approximation suitable. When $\eta \ll 1$, the derivative is overly smoothed, potentially suppressing meaningful dynamics. Conversely, $\eta \gg 1$ indicates excessive noise in the derivative, requiring additional smoothing or refined differentiation techniques.

The second metric evaluates how well the approximated derivative maintains consistency with the state transitions. Specifically, we compute the mean squared error (MSE) between the integrated derivative and the actual state difference over small time intervals:

\begin{equation}
    \rho = \frac{1}{N} \sum_{i=1}^{N} \left( x_i - \sum_{j=1}^{i} \dot{x}_j \Delta t \right)^2.
\end{equation}

This metric does not require knowledge of the true derivative but serves as a self-consistency check. By integrating the derivative, short-term fluctuations are smoothed, making long-term trends more apparent. A significant increase in this error in anomalous conditions suggests a breakdown in the causal approximation, indicating that the embedding dimension is insufficient to describe the system dynamics adequately.

\subsection{Anomaly Detection Logic}
We integrate the developed mathematical foundations into anomaly detection by leveraging state-derivative pairs in the latent space. Normal states are assumed to lie within a confined region, with deviations indicating anomalies.
Training uses the first 60\% of each engine's data, with an 90/10 split into training and validation sets.

Thresholds are computed as percentiles of the training data across all engines and a moving average is applied to all latent node values. Both are optimized for F1-score on anomalous last 40\% of the training dataset.  During testing, the model infers the derivative representation from a single time step, leveraging the fact that measurements at time $t$ inherently capture rate changes.

\textit{Detection logic:}
Each latent node's normal operating range is not predefined but dynamically recalibrated. Initially, the baseline is set using the first observed value. It is then updated using a moving average of the most recent 10 values, providing a real-time estimate of the normal distribution.
An instance is classified as anomalous (positive) if at least \emph{two} latent nodes simultaneously exceed their respective upper thresholds or fall below their lower thresholds; otherwise, it is considered normal (negative). 
Classification outcomes follow the standard notation: true negative (TN), false positive (FP), true positive (TP), and false negative (FN).
In high-stakes applications like aviation, failing to raise any alarm near end-of-life is unacceptable. In addition to conventional classification metrics, we introduce the \textbf{Critical Detection Rate (CDR)}, defined as the percentage of engines for which the detector produces at least one positive signal in the final 10\% of their lifetime. Although progressive degradation can lead to early false negatives that lower recall, these are less critical than missed detections near failure.

\section{Results}

\subsection{Consistency of the Proposed Approach}
First, we examine the consistency of the proposed approach. The loss terms consistently converge over 50 training epochs across five independent runs, indicating stable training behavior. Both the standard Mean Squared Error (MSE) loss and the newly introduced TDC loss term exhibit this convergence. For a visual representation, refer to Fig.~\ref{fig:CMPSS_mse_tdc_loss} in Appendix~\ref{sec:figures}. 
We conducted a Jacobian rank analysis using Singular Value Decomposition (SVD). Applying a zero-threshold of $\epsilon = 10^{-9}$, all samples in the test dataset exhibit a full-rank Jacobian. This result suggests that the mapping $\mathcal{E}$, which projects the measurement space $\mathbb{R}^k$ to an embedding space $\mathbb{R}^n$, behaves as an immersion, preserving the differential structure.\par

The first metric, $\sigma$, confirmed the validity of the central difference method for approximating the first time derivative, with values ranging from 0.7 to 1.2 across engines in the training split. The second metric assessed the correlation of state derivatives, capturing causal relationships under normal and anomalous conditions. Table~\ref{tab:rho_values} summarizes the mean $\pm$ standard deviation of the correlation coefficient $\rho$ in both cases.

\begin{table}[h]
    \centering
    \caption{Comparison of mean $\pm$ standard deviation values of $\rho$ for normal and anomalous conditions across latent node pairs.}
    \label{tab:rho_values}
    \begin{tabular}{lcc}
        \hline
        Latent Node Pair & Normal Condition ($\rho$) & Anomalous Condition ($\rho$) \\
        \hline
        (0-4) & $0.00037 \pm 0.00012$ & $0.00044 \pm 0.00016$ \\
        (1-5) & $0.00019 \pm 0.00002$ & $0.00034 \pm 0.00015$ \\
        (2-6) & $0.00026 \pm 0.00007$ & $0.00079 \pm 0.00027$ \\
        (3-7) & $0.00048 \pm 0.00014$ & $0.00073 \pm 0.00011$ \\
        \hline
    \end{tabular}
\end{table}

The results indicate that in the normal regime, the correlation values remain low with minimal variation. However, in the anomalous regime, we observe a slight increase in mean values and a broader standard deviation range, suggesting a stronger deviation in state derivative relationships under anomaly conditions.

\subsection{Phase Space Representation and State-Derivative Relationships}

\begin{figure}[h]
    \centering
    \includegraphics[width=\linewidth]{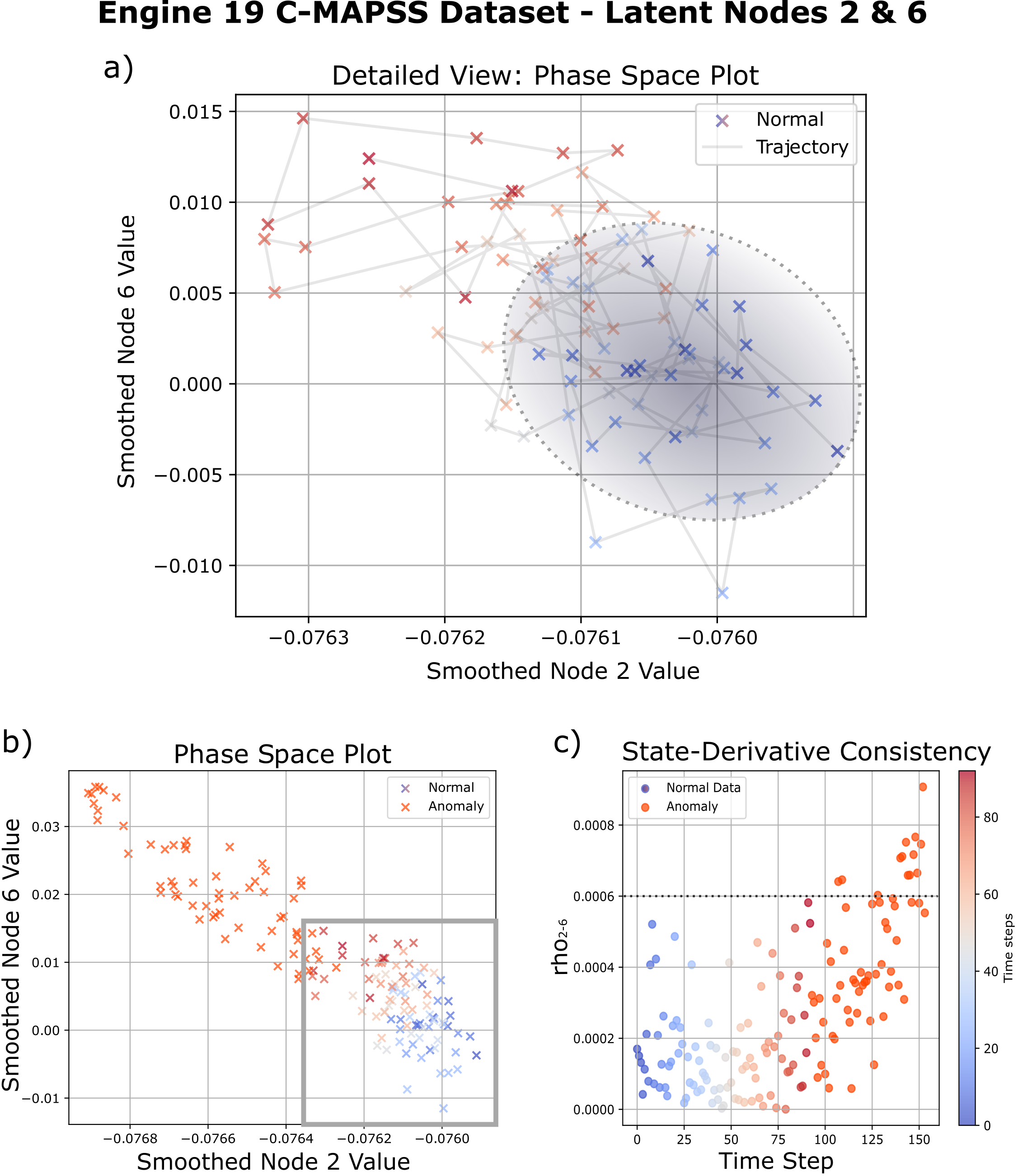}
    \caption{(a) A detailed view of (b) (highlighted grey box), showing phase-space trajectories and the attractor region at the beginning of the system dynamics. (b) Phase-space plot of the latent node pair (2, 6) average. (c) State-Derivative Consistency metric for this pair.
 The timestep color map applies to all figures.}
    \label{fig:CMAPSS_latent_detailed}
\end{figure}
\begin{figure*}[h]
    \centering
    \includegraphics[width=1.\linewidth]{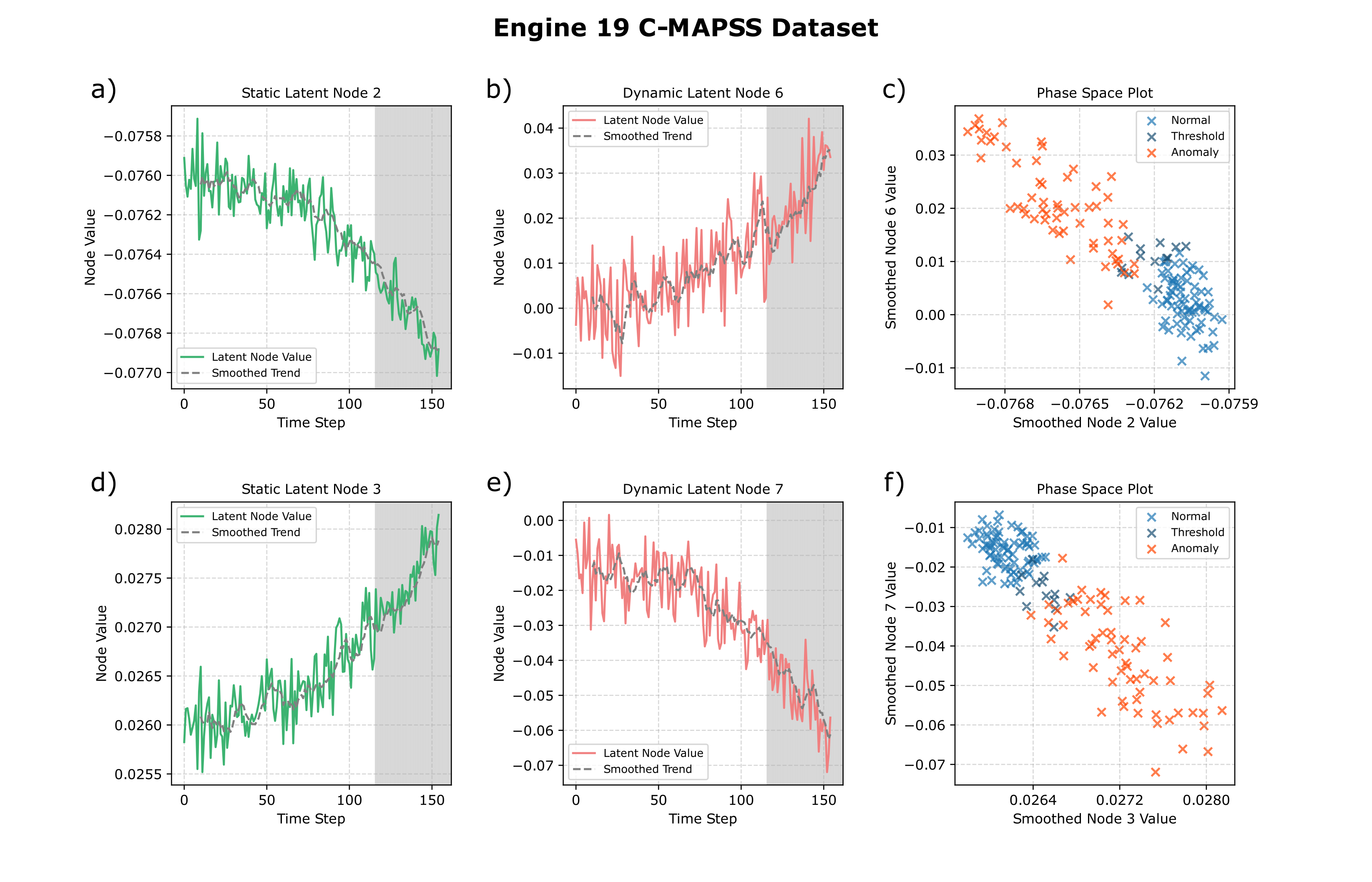}
    \caption{Latent node state–derivative pairs for nodes (2, 6) are shown in panels (a) and (b), and for nodes (3, 7) in panels (d) and (e). The corresponding phase-space representations for these pairs are displayed in panels (c) and (f), respectively. In all phase-space plots, the node values have been smoothed using a sliding average window of 3.}
    \label{fig:CMAPPS_latent_big}
\end{figure*}

Fig.~\ref{fig:CMAPPS_latent_big}a-f presents two state-derivative pairs along with their respective phase space representations. A gradual increase or decrease in both state and derivative nodes is observed as the system transitions toward the anomalous range. A gradual increase or decrease in both state and derivative nodes is observed as the system transitions toward the anomalous range. For the phase space plots, we slightly smoothed the node values using a moving average with a window of three. During the initial phase of the system dynamics, the states evolve more gradually, supporting the hypothesis that, in the normal regime, the dynamics follow a simple and stable attractor geometry. The phase space representation further indicates that normal states are embedded more densely than anomalies, with a continuous trend toward the anomalous region. \par

The Jacobian analysis suggests that the latent states are linearly independent, as indicated by the full-rank property observed in the Singular Value Decomposition (SVD) analysis. However, this does not inherently enforce an orthonormal projection, meaning that scale freedom exists and angular relationships are not necessarily preserved.

Nevertheless, from a phase-space perspective, the qualitative relationship between states remains intact. Anomalous states continue to map to regions outside the bounded domain of normal states, ensuring that the essential structural distinction is preserved, even if exact angles and scales are distorted.\par
Our objective is to develop methods that integrate physics-inspired consistency principles while imposing only the necessary constraints. This approach maintains the model’s flexibility and ability to learn effectively. Therefore, we do not enforce an explicit orthonormality condition in the AE design.
Let's have a more detailed view on the phase space evolution. Fig.~\ref{fig:CMAPSS_latent_detailed}a illustrates a clear drift toward the anomalous region, highlighting a gradual deviation in system behavior. Fig.~\ref{fig:CMAPSS_latent_detailed}c presents the evolution of $\rho$, which quantifies the causal approximation of state-derivative pairs. In the normal range, $\rho$ remains stable on average, indicating a consistent dynamical description. However, as the system enters the anomalous regime, a significant increase in $\rho$ is observed, suggesting a loss in the dynamical consistency of the system. The theoretical derivation showed that when system dynamics deviate from a stable state, the increasing complexity requires a higher-dimensional embedding to adequately capture the system’s behavior. We hypothesized that this effect would lead to the collapse of state-derivative approximations, serving as an indicator of anomalous behavior. Our observations now confirm this, as we see a loss of dynamical consistency in the latent space, reinforcing our hypothesis that anomalies emerge when the system transitions beyond the representational capacity of the learned embedding.

In Fig.~\ref{fig:CMAPSS_latent_detailed}b, we observe the evolution of the phase space representation over time, transitioning from normal to abnormal behavior. Initially, the phase space exhibits an attractor-like structure around which the dynamics evolve in a stable manner. As the system progresses, we observe increased dispersion and a gradual drift away from the attractor toward the anomalous region. While in the range characterized as normal, the state-derivative approximation remains stable, as seen in Fig.~\ref{fig:CMAPSS_latent_detailed}a. However, subtle deviations from the attractor’s dynamics can already be detected in the earliest phase of the transition. This aligns with our second hypothesized effect, which suggests that even minor disruptions in the system’s stability could serve as an early indicator of anomalies.\par

We numerically assess the injectivity of the encoder by evaluating pairwise distance ratios in latent space across more than 4,000 samples per subset. No instances were found that violated the injectivity criterion (i.e., minimum ratio below $10^{-5}$), indicating the absence of latent-space collisions.\par

Table~\ref{tab:subset_results} summarizes the performance of our model on the FD001 and FD003 subsets. We observe high overall accuracy, precision, recall, and F1-scores on both datasets, indicating consistent detection performance. The precision values are notably strong, and most false positives (FPs) occur near the boundary between the training (first 60\% of lifetime) and testing (last 40\%) regions, where the engine behavior starts to change. Similarly, the few false negatives (FNs) we observed are also concentrated around this transition region.
Importantly, the \textbf{Critical Detection Rate (CDR)}, the percentage of engines with at least one positive detection in the final 10\% of their lifetime, reaches 100\% on both datasets, demonstrating that the model successfully detects every failing engine before its end of life.

Please refer to Tab.~\ref{tab:hypParameterList} in Appendix~\ref{sec:Hyperparameter} for a detailed overview of the model architecture, training hyperparameters, and the threshold settings used for anomaly detection.

\begin{table}[ht]
\centering
\caption{Performance of TDC-AE across C-MAPSS subsets}
\scriptsize
\begin{tabular}{lccccc}
\toprule
\textbf{Dataset} & \textbf{Acc. (\%)} & \textbf{Prec. (\%)} & \textbf{Rec. (\%)} & \textbf{F1 (\%)} & \textbf{CDR (\%)} \\
\midrule
FD001 & 96.49 & 95.95 & 95.18 & 95.56 & 100.0 \\
FD003 & 94.84 & 93.81 & 93.16  & 93.49 & 100.0 \\

\bottomrule
\end{tabular}
\label{tab:subset_results}
\end{table}

\subsection{Comparison with Literature Benchmarks}
\begin{table*}[h]
\caption{Performance comparison of different approaches benchmarks for anomaly detection in the C-MAPSS dataset FD001.
}
\centering
\begin{tabularx}{0.7\textwidth}{lXXXXX}
\toprule
\textbf{Algorithm} & \textbf{Acc. (\%)} & \textbf{Prec. (\%)} & \textbf{Rec. (\%)} & \textbf{Spec.(\%)} & \textbf{F1 (\%)} \\
\midrule
\textbf{TDC-AE} & \textbf{96.49} & \textbf{95.95} &  95.18 &  97.35 & 95.56 \\ \hline
DeepLSTM-AE \cite{zhu_anomaly_2024} & 96.45 & 94.81 &  \textbf{98.12} & - &\textbf{96.44}\\  \hline
Random Forest \cite{zhu_anomaly_2024} & 92.33 & 97.40 & 86.66 & - & 91.71 \\  \hline
CNN-AE \cite{zhu_anomaly_2024} & 92.22 & 89.08 & 95.86 & - & 92.35 \\  \hline
SMOTE Transformer \cite{liu_deep_2023} & - & - & 92.22 & 93.44 & - \\ \hline
XGBoost \cite{zhu_anomaly_2024} & 84.57 & 77.73 & 96.00 & - & 85.90  \\ \hline
Dense-AE \cite{bataineh_autoencoder_2020} & - & 89.6 & 72.4 & - & 80.1 \\ 
\bottomrule
\end{tabularx}

\label{tab:performance_comparison}
\end{table*}
We have discussed the consistency of our method and analyzed how the TDC-AE aligns with the mathematical foundation developed in this paper. Now, we evaluate its detection performance in comparison to literature benchmarks.

Tab.~\ref{tab:performance_comparison} presents a comparison across different architectures. Our approach achieves competitive results across all detection metrics. While more complex models such as LSTM-AEs and Transformers are designed to capture long-range temporal dependencies, our method attains comparable performance with a substantially simpler architecture. In \cite{zhu_anomaly_2024}, the exact LSTM architecture is not explicitly provided, so we base our assumptions on similar studies of complex physical systems \cite{mahmoud_ae-lstm_2022}, considering the following configuration: sequence length \( L = 48 \), hidden dimension \( d_{\text{hidden}} = 16 \), and input dimension \( d_{\text{input}} = 24 \). This results in a Multiply-Accumulate Operations (MAC) count of 245,760.

In contrast, our approach achieves superior detection performance with significantly lower computational complexity, requiring only about 3360 MACs. 

Furthermore, a variant of the TDC-based architecture has also been applied to the BATADAL dataset, which models cyber-physical water distribution systems. This approach outperformed the challenge-winning method \cite{housh_model-based_2018}, which is grounded in explicit physical modeling. These results underscore the potential of the proposed framework to generalize across structurally different dynamical systems and application domains \cite{somma_hybrid_2025}.

\section{Conclusion}
We introduced an unsupervised framework for anomaly detection in complex physical systems, grounded in \textbf{classical embedding theory} and physics-inspired consistency principles, particularly \textbf{state-derivative relations} in the embedding space. To translate this theoretical foundation into practice, we developed \textbf{TDC-AE}, an algorithm designed to capture system dynamics efficiently.

The results demonstrate that our embedding strategy can drastically reduce model size. On the C-MAPSS dataset, TDC-AE matches benchmark performance while achieving nearly a 100× reduction in MAC operations compared to LSTM models, bringing the computation down to about 3,000 MACs. This efficiency makes our approach particularly well-suited for lightweight edge computing.

As a next step, we plan to investigate the theoretical implications of having access to a system-level state representation, effectively a phase plane of complex systems. Exploring this higher-level view could provide new insights into the underlying dynamics, stability properties, and fault evolution mechanisms, enabling more principled approaches to monitoring and control

\appendix
\section{Theoretical Background} \label{sec:definitions_implementation}

\subsection{General Definition of an Oscillatory Dynamical System}
\begin{figure}[h]
    \centering
    \includegraphics[width=0.75\linewidth]{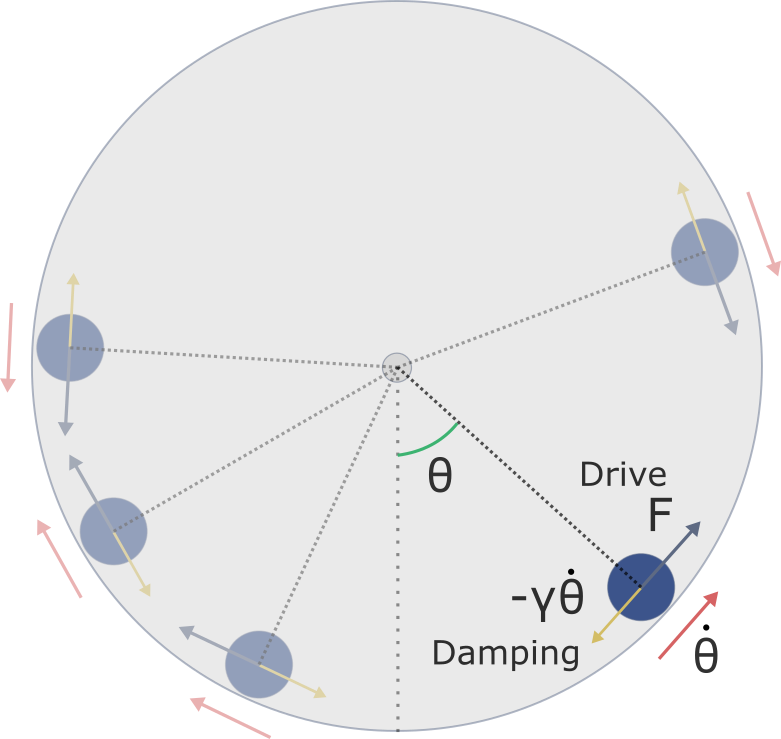}
    \caption{A driven damped pendulum with angular displacement $\theta$, velocity $\dot{\theta}$, damping $-\gamma \dot{\theta}$, and external driving force $F$.}

    \label{fig:CMPSS_pendulum_scheme}
\end{figure}
A general oscillatory dynamical system is governed by a set of differential equations describing periodic motion. In its most general form, the system evolves according to
\begin{equation}
    \dot{\mathbf{x}} = \mathbf{f}(\mathbf{x}, t),
\end{equation}
where $\mathbf{x} \in \mathbb{R}^n$ represents the state variables, and $\mathbf{f}(\mathbf{x}, t)$ defines the governing dynamics. A common class of oscillatory systems follows a second-order differential equation of the form
\begin{equation}
    \ddot{\theta} + g(\theta, \dot{\theta}) + h(\theta, t) = 0,
\end{equation}
where $g(\theta, \dot{\theta})$ represents dissipative forces, such as damping, and $h(\theta, t)$ accounts for external periodic forcing. A widely studied case is the damped driven oscillator, as shown in Fig.~\ref{fig:CMPSS_pendulum_scheme}, which satisfies the equation
\begin{equation}
    \ddot{\theta} + \gamma \dot{\theta} + \omega_0^2 \sin\theta = A \cos(\omega_{\text{drive}} t),
\end{equation}
where $\theta$ denotes the angular displacement and $\dot{\theta}$ its corresponding angular velocity. The parameters $\gamma$ and $\omega_0$ represent the damping coefficient and the natural frequency of the system, respectively, while $A$ and $\omega_{\text{drive}}$ define the amplitude and frequency of an external periodic driving force \cite{goldstein_classical_2000}. To analyze deviations from the nominal oscillatory state, we introduce a slow drift in both the angular displacement and velocity components by modifying the evolution equations as
\begin{equation}
    {\theta} \leftarrow {\theta} + \alpha t, \quad
    \dot{\theta} \leftarrow \dot{\theta} + \beta t.
\end{equation}

For illustration, we consider a system with parameters $\gamma = 0.2$, $\omega_0 = 1.0$, $A = 0.8$, $\omega_{\text{drive}} = 1.2$, and drift rates $\alpha = 0.005$ and $\beta = 0.002$. Under these conditions, the system initially remains within a bounded oscillatory regime but gradually exhibits deviations due to the drift terms, leading to an eventual departure from the confined state space. The effect of this perturbation is visualized in Figure~\ref{fig:CMPSS_pendulum_theta_phasespac}, where the phase space trajectory initially remains within a bounded region but progressively transitions into an unbounded state. 

\begin{definition}[Box-Counting Dimension {\cite{falconer_fractal_1990}}] 
\label{def:box_counting_dim}
Let \( A \) be a bounded subset of \( \mathbb{R}^k \). For each \( \epsilon > 0 \), let \( N_{\epsilon}(A) \) denote the smallest number of sets of diameter at most \( \epsilon \) required to cover \( A \).  

The \emph{box-counting dimension} of \( A \) is defined as
\[
\dim_B(A) = \lim_{\epsilon \to 0} \frac{\ln N_{\epsilon}(A)}{\ln(1/\epsilon)}
\]
provided the limit exists.
\end{definition}

\begin{theorem}[Fractal Whitney Embedding Prevalence Theorem {\cite{sauer_embedology_1991}}]
\label{th:fractal_emeddbing}
Let \( A \) be a compact subset of \( \mathbb{R}^k \) with box-counting dimension \( d \), and let \( n \) be an integer greater than \( 2d \). For almost every smooth map \( F: \mathbb{R}^k \to \mathbb{R}^n \),
\begin{enumerate}
    \item \( F \) is one-to-one on \( A \).
    \item \( F \) is an immersion on each compact subset \( C \) of a smooth manifold contained in \( A \).
\end{enumerate}
\end{theorem}

\section{Implementation details and numerical methods}\label{sec:Implementation}

\subsection{The Box-Counting Dimension in a dammped driven Pendulum}

To illustrate the concept of fractal dimension, we numerically estimated the \textit{box-counting dimension} for a \textit{slice of the phase space} of the damped driven pendulum. We considered the first \textbf{100–1000} time steps and applied the box-counting method to a subset of the \((\theta, \dot{\omega})\) plane.

Using a range of box sizes \(\epsilon\), we counted the number of occupied boxes \(N(\epsilon)\) that contained at least one trajectory point. The chosen values of \(\epsilon\) were logarithmically spaced as:
\[
\epsilon \in \{10^{-2}, 10^{-1.9}, 10^{-1.8}, \dots, 10^{-0.4} \}.
\]
These values were selected to balance resolution constraints and ensure a meaningful estimation within the given simulation timestep.

The relationship between \(\log N(\epsilon)\) and \(\log(1/\epsilon)\) was then analyzed through a \textbf{linear fit}, where the slope provides an estimate of the \textit{box-counting dimension}. 

\begin{figure}
    \centering
    \includegraphics[width=\linewidth]{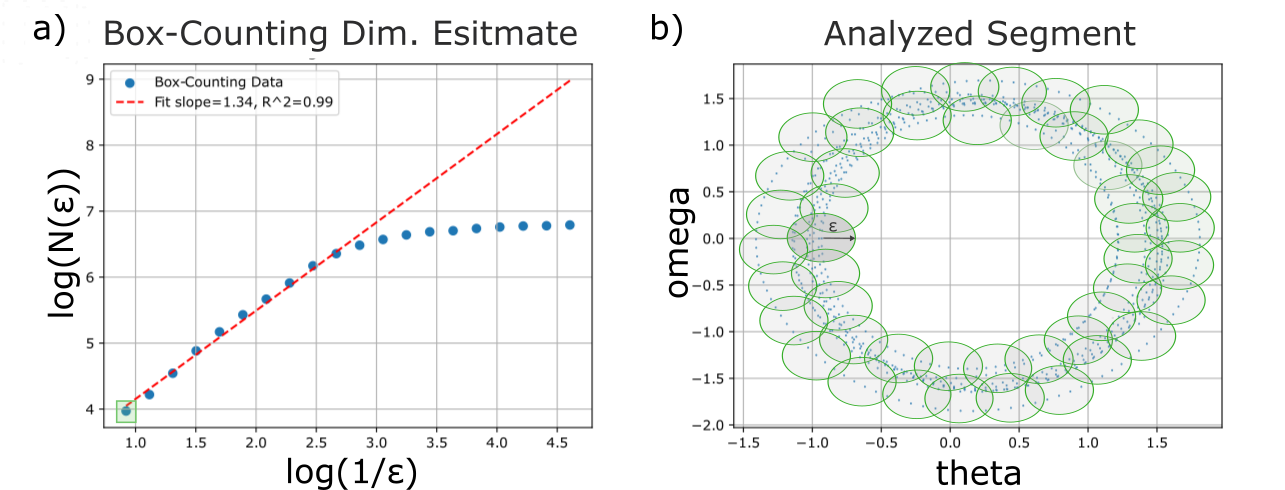}
    \caption{(a) Log-log plot of the box-counting method, showing the linear fit used to estimate the fractal dimension.  
(b) Phase-space slice of the damped driven pendulum used for the computation. The schematic shows 48 grid cells with $\epsilon \approx 0.45$. The roughly corresponding value pair is highlighted as a green square in panel (a).}
    \label{fig:box_count_pendulum}
\end{figure}

Fig.~\ref{fig:box_count_pendulum}a illustrates the fitting result, and Fig.~\ref{fig:box_count_pendulum}b shows the selected phase-space slice used for the computation.

\subsection{TDC-AE}
\begin{algorithm}
\label{alg:TDC_informed_training}
\caption{Temporal Differential Consistency Loss Enhanced Autoencoder (TDC-AE)}
\begin{algorithmic}[1]
\STATE \textbf{Initialize} Autoencoder with input \( X_t \in \mathbb{R}^k \) and latent variables and time derivatives (\( \mathbf{z} ,\dot{\mathbf{z}}) \in \mathbb{R}^n \), Optimizer, MSE Loss
\FOR{each epoch in training\_epochs}
    \FOR{each batch in training\_data}
        \STATE Perform forward pass through the Autoencoder and Encoder: \\
        \( X_{\text{rec}} \gets \text{Autoencoder}(X_t) \)
        \STATE Compute latent representations for neighboring time steps: \\
        \( (\mathbf{z}_{t-1}, \dot{\mathbf{z}}_{t-1})   \gets \text{Encoder}(X_{t-1}) \) \\
        \( (\mathbf{z}_{t+1}, \dot{\mathbf{z}}_{t+1})   \gets \text{Encoder}(X_{t+1}) \) 
        \STATE Compute central difference derivative: \\
        \( \Delta_t \mathbf{z} \gets ( \mathbf{z}_{t+1} - \mathbf{z}_{t-1}) / 2\Delta_t \)
        \STATE Compute temporal differential consistency loss using MSE:\\ 
        \( \text{TDC-Loss} \gets \text{MSE}(\Delta_t \mathbf{z}, \dot{\mathbf{z}}) \)
        \STATE Compute reconstruction loss using MSE: \\
        \( \text{Rec-Loss} \gets \text{MSE}(X_{\text{rec}}, X_t) \)
        \STATE Backpropagate the total loss: \\
        Compute gradients w.r.t. model parameters for \( \text{Rec-Loss} + \alpha \cdot \text{TDC-Loss} \)
        \STATE Update the Autoencoder parameters using the optimizer
    \ENDFOR
\ENDFOR
\end{algorithmic}
\end{algorithm}
\subsection{Implementation of the TDC-AE and Hardware} \label{sec:Hyperparameter}

\begin{table}[h]
\centering
\caption{Hyperparameters used in training the autoencoder and derived thresholds for the C-MAPSS (FD001 and FD003) datasets.}
\label{tab:hypParameterList}
\begin{tabular}{p{3cm} p{5cm}}
\toprule
\textbf{Hyperparameter} & \textbf{Value} \\
\midrule
Optimizer & Adamax \\
Activation Function & Tanh \\
Number of Dense Layers & 6 \\
Architecture & 24-24-24-8-24-24-24 \\
Learning Rate & 0.003 \\
Batch Size & 32 \\
$\alpha$ & 100 \\
Threshold (FD001) & max = 86th perc., min = 9th perc. \\
Threshold (FD003) & max = 75th perc., min = 22nd perc. \\
Moving Average Window & 12 \\
\bottomrule
\end{tabular}
\end{table}

The experiments were conducted on a system running Windows 11, powered by a 13th Gen Intel Core i7-13700H processor (2.4 GHz, 14 cores, 20 threads), along with 32 GB of RAM and an NVIDIA GeForce RTX 4070 Laptop GPU.

The model was implemented using PyTorch \textit{2.3.0} with CUDA \textit{11.8}, leveraging GPU acceleration to enhance computational efficiency and reduce training time. The autoencoder's architecture and training setup are detailed in Table~\ref{tab:hypParameterList}. 
\section{Figures} \label{sec:figures}

\begin{figure}[h]
    \centering
    \includegraphics[width=\linewidth]{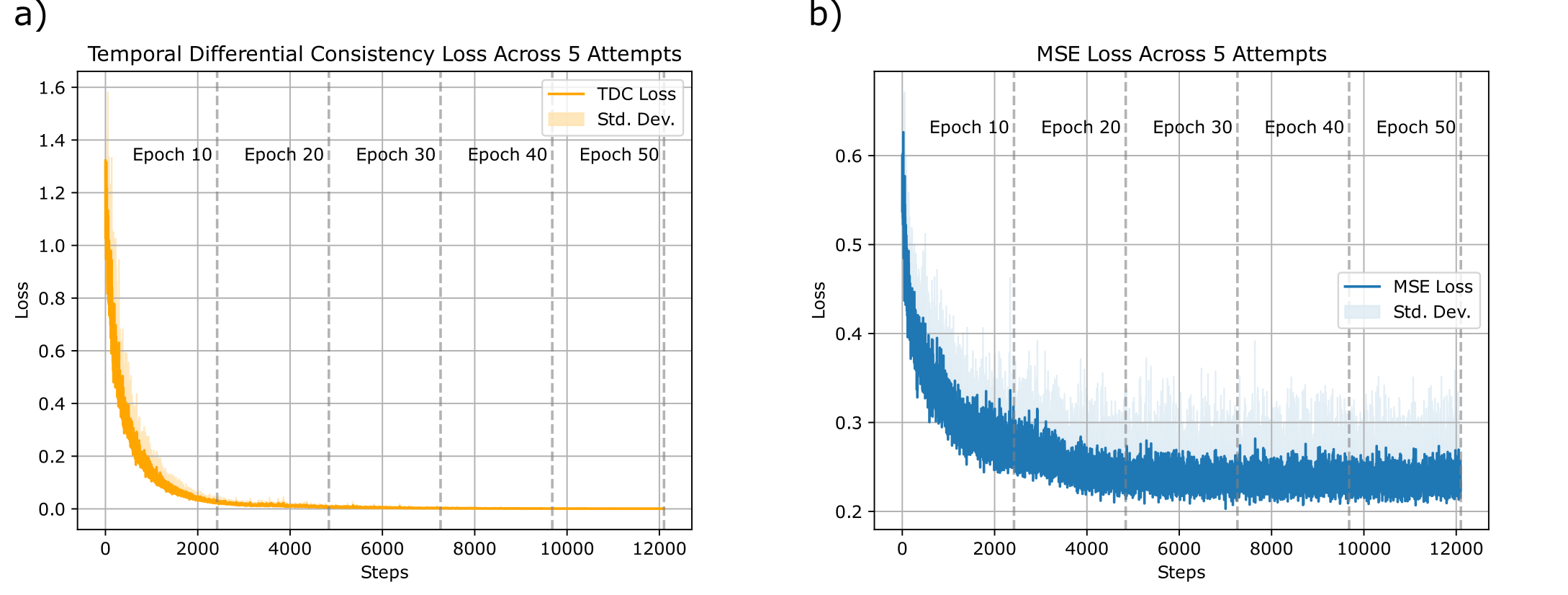}
    \caption{The two loss terms used in TDC-informed training across five independent attempts on the C-MAPSS test dataset.}
    \label{fig:CMPSS_mse_tdc_loss}
\end{figure}

\begin{figure}[h]
    \centering
    \includegraphics[width=0.8\linewidth]{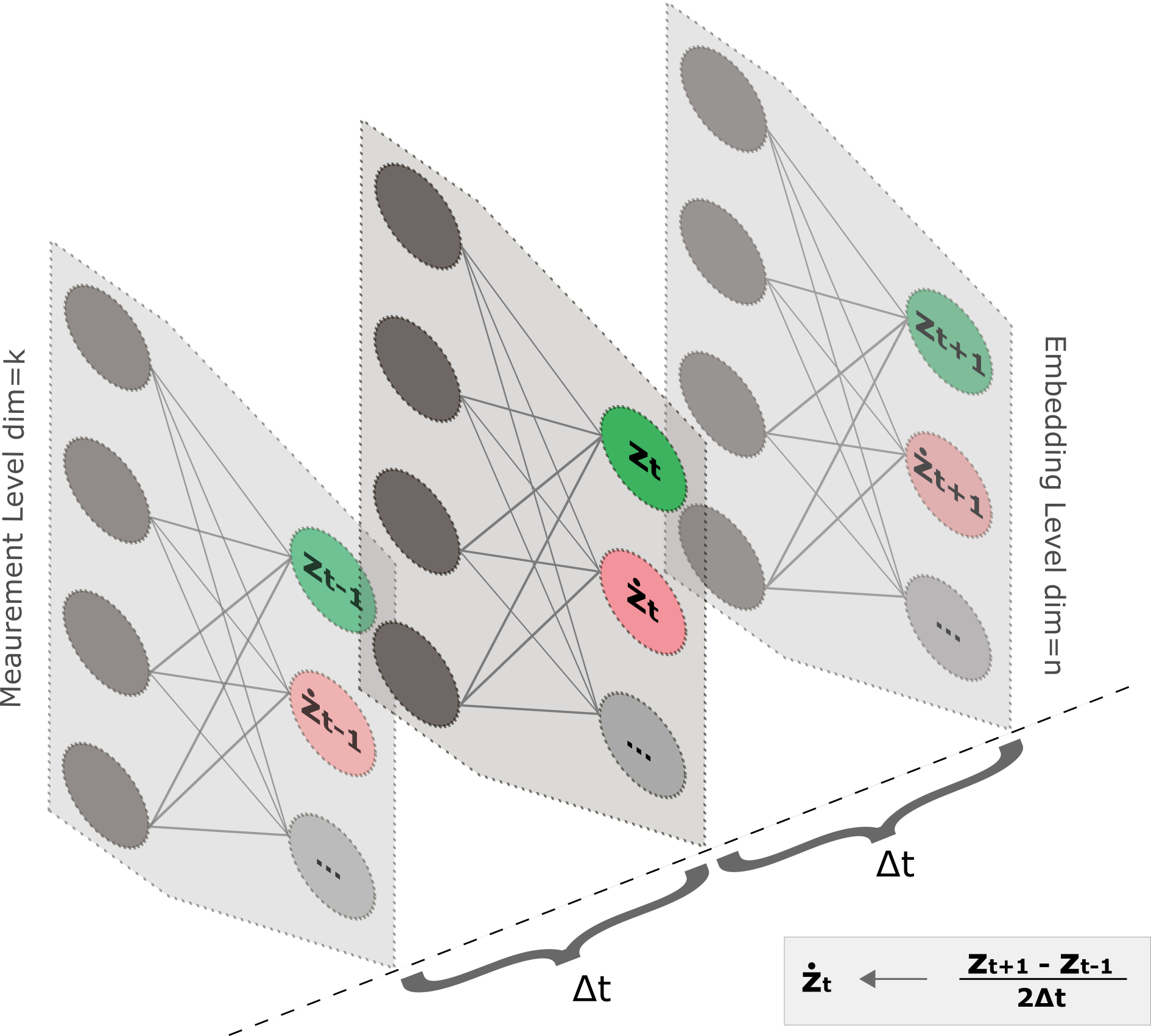}
    \caption{Schematic of temporal difference consistency in the latent space of an autoencoder. The derivative node $z_{\dot{}}$ is aligned with the central difference of the static node $z$ using time steps $t-1$, $t$, and $t+1$.}
    \label{fig:TDC_AE_scheme}
\end{figure}


\end{document}